\documentclass[letterpaper]{article} 
\usepackage{aaai25}  
\usepackage{times}  
\usepackage{helvet}  
\usepackage{courier}  
\usepackage[hyphens]{url}  
\usepackage{graphicx} 
\usepackage{natbib}  
\usepackage{caption} 
\usepackage{algorithm}
\usepackage{algorithmic}

\usepackage{multirow}

\usepackage{newfloat}
\usepackage{listings}
\DeclareCaptionStyle{ruled}{labelfont=normalfont,labelsep=colon,strut=off} 
\floatstyle{ruled}
\newfloat{listing}{tb}{lst}{}
\floatname{listing}{Listing}
\title{SR-FoT: A Syllogistic-Reasoning Framework of Thought for Large Language Models Tackling Knowledge-based Reasoning Tasks}
\author {
    Wentao Wan\textsuperscript{\rm 1},
    Zhuojie Yang\textsuperscript{\rm 1},
    Yongcan Chen\textsuperscript{\rm 3},
    Chenglin Luo\textsuperscript{\rm 1},
    Ruilin Wang\textsuperscript{\rm 1},
    Kehao Cai\textsuperscript{\rm 1},
    Nan Kang\textsuperscript{\rm 1},
    Liang Lin\textsuperscript{\rm 1},
    Keze Wang\textsuperscript{\rm 1,2}\thanks{Corresponding author.}
}
\affiliations {
    \textsuperscript{\rm 1}School of Computer Science and Engineering, Sun Yat-sen University\\
    \textsuperscript{\rm 2}Guangdong Key Laboratory of Big Data Analysis and Processing\\
    \textsuperscript{\rm 3}South China Normal University\\
    \{wanwt3, yangzhj53\}@mail2.sysu.edu.cn, bebobi@163.com, chenglinluo@foxmail.com, wangrlin23@mail2.sysu.edu.cn, caikehao@triody.tech, \{kangn5@mail2, linlng@mail\}.sysu.edu.cn, kezewang@gmail.com
}

\usepackage{bibentry}

\begin{document}

\maketitle

\begin{abstract}
Deductive reasoning is a crucial logical capability that assists us in solving complex problems based on existing knowledge. Although augmented by Chain-of-Thought prompts, Large Language Models (LLMs) might not follow the correct reasoning paths. 
Enhancing the deductive reasoning abilities of LLMs, and leveraging their extensive built-in knowledge for various reasoning tasks, remains an open question. Attempting to mimic the human deductive reasoning paradigm, we propose a multi-stage \textbf{S}yllogistic-\textbf{R}easoning \textbf{F}ramework \textbf{o}f \textbf{T}hought (\textbf{SR-FoT}) that enables LLMs to perform syllogistic deductive reasoning to handle complex knowledge-based reasoning tasks. 
Our SR-FoT begins by interpreting the question and then uses the interpretation and the original question to propose a suitable major premise. It proceeds by generating and answering minor premise questions in two stages to match the minor premises. Finally, it guides LLMs to use the previously generated major and minor premises to perform syllogistic deductive reasoning to derive the answer to the original question. 
Extensive and thorough experiments on knowledge-based reasoning tasks have demonstrated the effectiveness and advantages of our SR-FoT.
\begin{links}
\link{Code}{https://github.com/RodeWayne/SR-FoT}
\end{links}

\end{abstract}

\section{Introduction}
Deductive reasoning~\cite{johnson1999deductive} is the process of drawing valid inferences. 
Deductive reasoning is a powerful human capability, where rigorous deductive reasoning helps us use existing knowledge as premises to derive correct subsequent conclusions, enabling us to tackle various complex tasks in the real world.

Automated deductive reasoning has long been a pursuit in the field of Natural Language Processing (NLP)~\cite{chowdhary2020natural, bharadiya2023comprehensive, khurana2023natural}. Works on automated rigorous reasoning include reasoning engines and Automated Theorem Proving (ATP)~\cite{bibel2013automated}, which often provide methods for automatically checking the rigor of reasoning. However, these engines require the use of formal languages, which limits their applicability in knowledge-based reasoning scenarios. 
Because formal language-based reasoning requires a predefined library of formalized premises, many knowledge-based reasoning tasks, including common-sense question answering, involve a diverse array of premises. It is difficult to prepare and rigorously formalize such a large library of premises in advance.
Therefore, performing correct deductive reasoning in natural language holds significant importance.

Large Language Models (LLMs)~\cite{chang2024survey, floridi2020gpt, touvron2023llama, vicuna2023, huang2022towards, deepseekv2} pre-trained on extensive corpora possess inherent soft deductive reasoning capabilities~\cite{seals2023evaluating}. With the aid of the Chain-of-Thought prompt (CoT)~\cite{lyu2023faithful, wei2022chain, zhang2022automatic, turpin2024language, lee2024applying, liu2023federated}, the cognitive abilities of LLMs are further enhanced. However, reasoning with CoT often does not constitute strict deductive reasoning and thus can lack rigor. 
Fig.~\ref{nlp_case2} 
illustrates the different processes of handling the same problem using CoT and classic syllogistic deductive reasoning, clearly showing that the syllogistic deductive approach is more rigorous. We believe that guiding large language models to perform deductive reasoning, rather than merely multi-step reasoning, can enhance the rigor of the reasoning process, reduce illusions, and subsequently improve performance on complex tasks.

Inspired by the most fundamental human deductive reasoning paradigm, syllogistic reasoning~\cite{bucciarelli1999strategies, khemlani2012theories, bara1995development}, we propose a multi-stage reasoning framework for large language models to guide them in using syllogistic reasoning to solve specific problems. In contrast to existing works in the community~\cite{wu2023hence, ye2023assessing, deng2023syllogistic}, we do not solely rely on simplistic processes or create targeted benchmarks to evaluate the LLMs' capabilities in performing syllogistic reasoning. Instead, we propose a reasoning framework based on the syllogistic thinking paradigm to handle complex knowledge-based reasoning tasks. Our SR-FoT advances in guiding LLMs in performing syllogistic deductive reasoning, thereby achieving improved performance on these tasks and enhancing the rigor and reliability of the reasoning process.

Our SR-FoT consists of five stages. Initially, it involves interpreting the question. Subsequently, 
SR-FoT guides the Large Language Model (LLM) proposing a major premise suitable for the question. This major premise typically encompasses the built-in knowledge of the LLM, which serves as a universal rule that aids in addressing the original question. The next stage involves obtaining a minor premise, which acts as the bridge linking the major premise to the original problem and is crucial for applying the major premise to the current issue. 
We first let the LLM formulate minor premise questions based on the original question, major premise, and contextual information, and then answer these to obtain an appropriate minor premise. 
Finally, with both the major and minor premises established, we enable the LLM to perform syllogistic reasoning based on the original question and these premises to derive the answer to the original question. Furthermore, to minimize the interference caused by excessive information during the reasoning process~\cite{dong2023revisit}, we restrict each stage to only access the content from its necessary preceding stages. For example, during the final syllogistic reasoning, only the original problem and the previously established major and minor premises are visible, without the need to reference the problem interpretation and minor premise question stages.

Our \textbf{main} contributions can be summarized in three points: i) We propose a multi-step thinking framework that guides LLMs in using syllogistic deductive reasoning to solve knowledge-based reasoning tasks. Specifically, to enhance the ability of LLM to leverage its built-in knowledge for solving diverse tasks, we introduce a problem interpretation stage when acquiring the major premise and improve the quality of both premises as well as their logical connection to the original problem by adopting an autonomous question-answering approach during the acquisition of the minor premise~\cite{bubeck2023sparks}; ii) To facilitate more rigorous reasoning, we have designed our thinking framework so that each step only accesses the information necessary for that stage, thereby reducing the illusions and error accumulation that can come from overly long reasoning steps; iii) Our SR-FoT achieves superior performance over the existing chain of thought-related methods on various knowledge-based reasoning QA datasets such as ScienceQA~\cite{lu2022learn}, StrategyQA~\cite{geva2021did}, and BoolQ~\cite{clark2019boolq}, demonstrating the superiority of our SR-FoT.

\section{Related Work}
\subsection{Chain-of-Thought}
Chain-of-Thought~\cite{wei2022chain} has been demonstrated to enhance performance in reasoning tasks by fully utilizing the in-context learning capability of the large language model to stimulate its multi-step reasoning ability. Self-consistency CoT (SC-CoT)~\cite{wang2022self} further improves the performance of CoT by utilizing the consistency of multiple sampled reasoning chains. Complexity-based CoT (C-CoT)~\cite{fu2022complexity} further discovers that the consistency of complex reasoning chains is even more vital for the reasoning performance of language models. In addition, some efforts have also been made to further stimulate the reasoning ability of language models by focusing on the structure of the reasoning chain and the levels of reasoning, such as Least-to-Most~\cite{zhou2022least} and Tree-of-Thought~\cite{yao2023tree}. However, these works have not considered how to stimulate the reasoning abilities of LLMs from the perspective of logical reasoning.

\subsection{Logical Reasoning Ability of LLMs}
There has been considerable research within the community on the logical reasoning capabilities of LLMs, broadly categorized into two directions: one focuses on logic reasoning based on formal languages, and the other on natural language logic reasoning. Research related to formal language-based logic reasoning primarily concentrates on the field of Automated Theorem Proving (ATP)~\cite{bibel2013automated}, utilizing the built-in mathematical priors of LLMs to accelerate the search process in theorem proving or to construct a growing library of mathematical theorems to aid new proofs~\cite{wang2023legoprover}. This work typically operates within interactive theorem-proving platforms like the Lean system, which restricts its application in daily question-answering scenarios. Logic reasoning on natural language with LLMs generally involves soft reasoning~\cite{yu2023natural}, which does not provide rigorous guarantees. For instance, the Chain-of-Thought (CoT) enhances the general explicit reasoning abilities of LLMs, and there are exploratory studies demonstrating to what extent LLMs can perform in logical reasoning, or how segment checking might reduce soft deductive reasoning illusions and error accumulation~\cite{ye2023cognitive, dhuliawala2023chain}. Recently, several studies on syllogistic reasoning with LLMs have been proposed. However, these primarily create benchmarks~\cite{ye2023assessing}, evaluating the capability of LLMs to perform syllogistic reasoning on datasets with given premises. Unlike previous works, our research investigates how to guide LLMs through a multi-stage process that involves autonomously generating minor and major premises and performing syllogistic deductive reasoning to answer a variety of knowledge-based reasoning questions. 

\section{Methodology}
We have designed a reasoning framework that guides large language models to perform syllogistic deductive reasoning for addressing various knowledge-based reasoning question-answer tasks. Next, we present syllogistic reasoning as background knowledge, followed by a detailed description of our SR-FoT framework.
\subsection{Background: Syllogism}
\begin{figure}[t]
  \includegraphics[width=\columnwidth, trim={2cm 13cm 17cm 2cm}, clip]{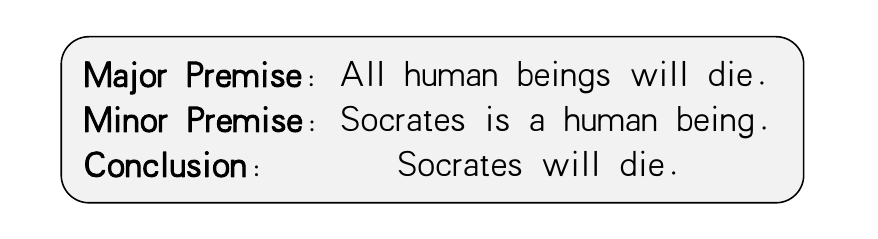}
  \caption{A syllogism example.}
  \label{fig: Syllogism example}
\end{figure}

In traditional logic, syllogism~\cite{smiley1973syllogism} is a form of reasoning where one proposition (the conclusion) necessarily follows from two other propositions (known as premises). As shown in Fig.~\ref{fig: Syllogism example}, a syllogism consists of three parts: a major premise, a minor premise, and a conclusion. Logically, the conclusion is derived by applying the major premise to the minor premise. The major premise represents a general principle, while the minor premise is a specific statement. Syllogistic reasoning is a type of deductive reasoning; rigorous deductive reasoning ensures that if the premises are correct, the conclusion must also be correct.

\subsection{Procedure of Our SR-FoT}
\begin{figure*}[t]
  \centering
  \includegraphics[width=\textwidth]{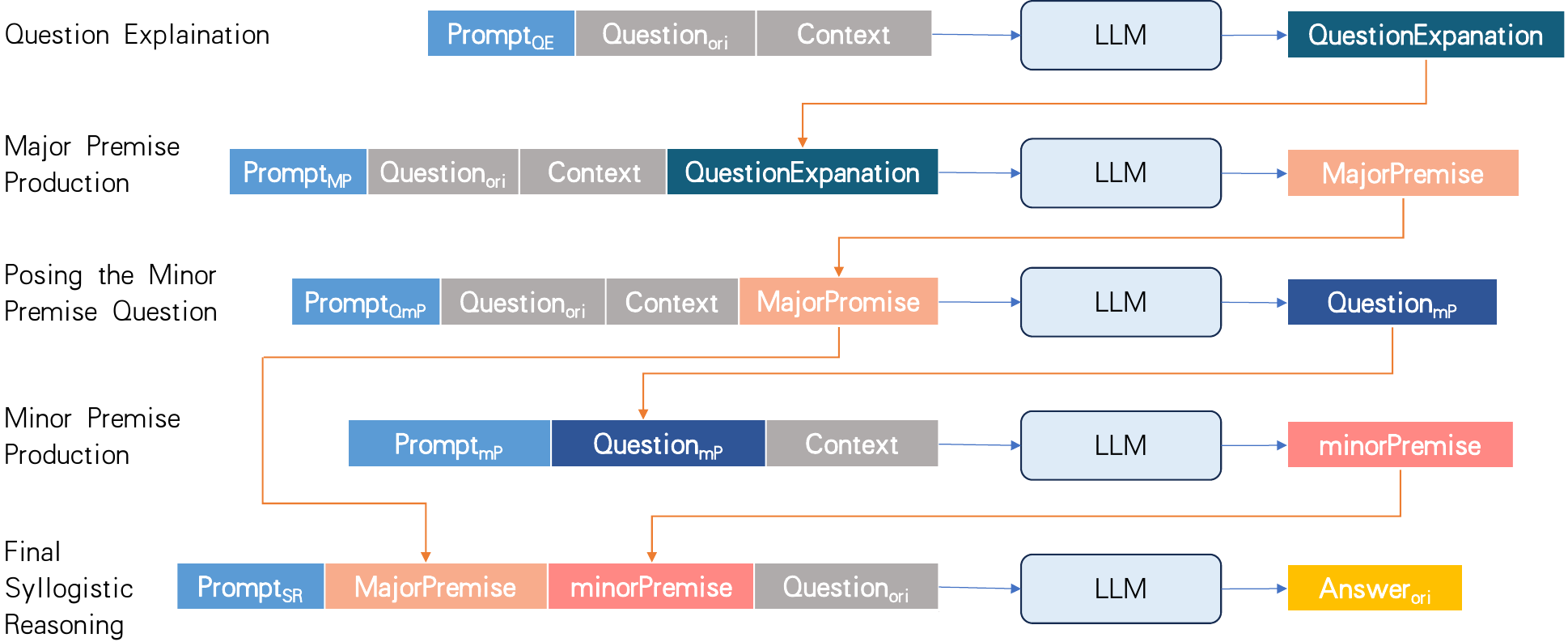}
  \caption{Procedure of our SR-FoT. Question\textsubscript{ori}: Original Question, Context: Context provided for Original Question, Answer\textsubscript{ori}: Answer for Original Question, Question\textsubscript{mP}: Question for Minor Premise, Propmt\textsubscript{CoT}: Guide Prompt for CoT, Propmt\textsubscript{QE}: Guide Prompt for Question Explanation, Propmt\textsubscript{MP}: Guide Prompt for Major Premise Production, Propmt\textsubscript{QmP}: Guide Prompt for Posing the Minor Premise Question, Prompt\textsubscript{mP}: Guide Prompt for Minor Premise Production, Prompt\textsubscript{SR}: Guide Prompt for Final Syllogistic Reasoning and so on.}
  \label{fig: SR-Fot pipeline}
\end{figure*}

\begin{figure*}[]
  \centering
  \includegraphics[width=\textwidth]{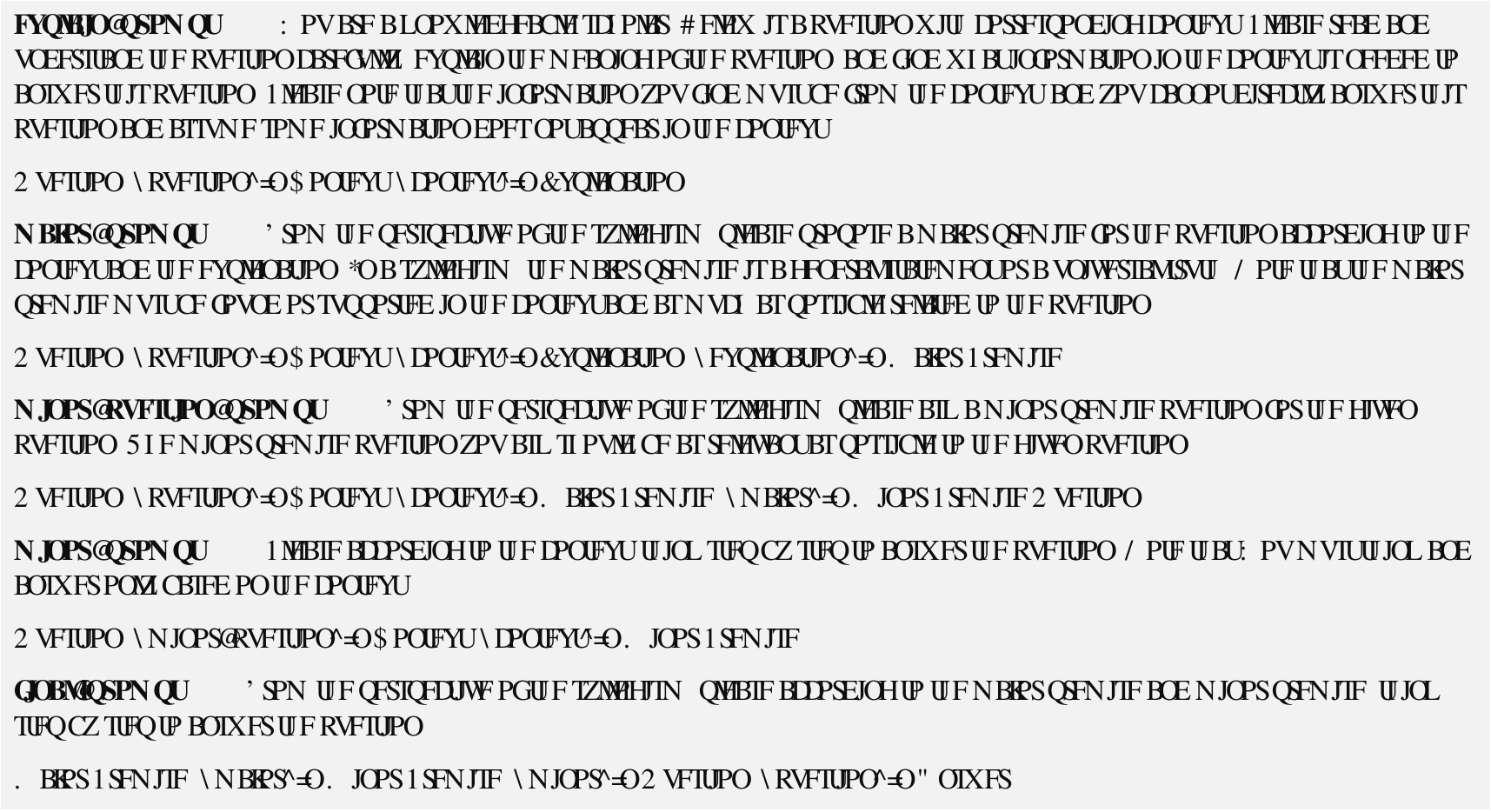}
  \caption{Prompts for each stage of our SR-FoT.}
  \label{fig: SR-FoT_prompt}
\end{figure*}

While our proposed SR-FoT does not guarantee the execution of rigorous syllogistic reasoning for 100\%, it aims to guide the reasoning paradigms of the LLM through carefully designed prompts and sub-tasks at each stage. By strategically controlling the input visible at each stage, we strive to ensure that the LLM conducts rigorous syllogistic reasoning and minimizes the occurrence of reasoning fallacies. Specifically, As shown in Fig.~\ref{fig: SR-Fot pipeline}, our SR-FoT is divided into five stages. The prompts for each stage of our SR-FoT are shown in Fig.~\ref{fig: SR-FoT_prompt}.
\begin{figure*}[h]
\centering
\includegraphics[width=1\textwidth]{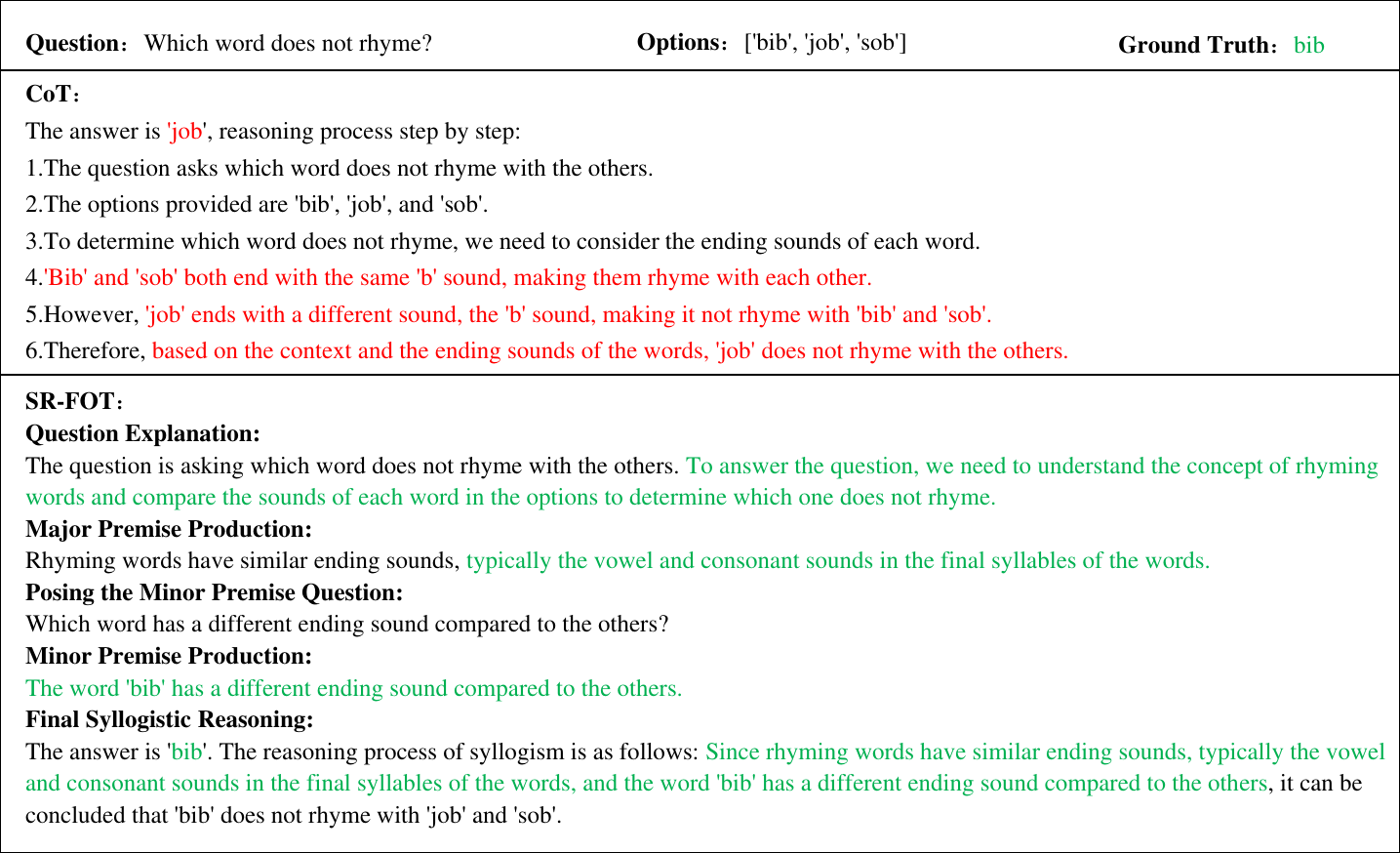}
\vspace{-12pt}
\caption{A case of using CoT and SR-FoT to answer a question in the ScienceQA dataset respectively. The highlighted red parts indicate the incorrect or misleading content, while the highlighted green parts indicate the content that helps correct reasoning.}
\vspace{-10pt}
\label{nlp_case2}
\end{figure*}

Stage 1: \textbf{Question Explanation}. 
The key to utilizing syllogistic reasoning to solve various complex knowledge-based reasoning tasks lies in formulating appropriate major and minor premises that fit the current problem. Accordingly, the first stage of our SR-FoT involves using a prompt with examples to guide the LLM in interpreting the original task question and proposing a solution approach. This guidance helps direct the LLM to formulate suitable major premises and then appropriate minor ones. In this stage, besides the guidance and example prompts, the only information available to the LLM is the "original question" and the "context" provided by the task, which also includes "options" information for multiple-choice questions.

Stage 2: \textbf{Major Premise Production}.
After acquiring the ``question explanation'', we gain a deeper understanding of the original question and develop an approach to solve it. This solution approach often includes guidance on what further information is needed. In this stage, based on these guidelines, we propose an appropriate major premise, which is derived from the task "context" or the built-in knowledge of the LLM. In Stage 2, besides the guidance and example prompts, the information accessible to the LLM includes the "original question" "context" and the "question explanation" generated in the first stage.

Stage 3: \textbf{Posing the Minor Premise Question}.
After establishing the major premise, to effectively engage in syllogistic reasoning, we need a minor premise. In syllogistic reasoning, the minor premise is a specific statement that describes the relationship between a particular instance and the category mentioned in the major premise. Through the minor premise, the universal characteristics of the major premise can be applied to the specific instance in the minor premise, which is a crucial step in using syllogistic reasoning to solve specific problems. Given the diverse and often complex nature of the knowledge-based reasoning tasks we need to address, it is challenging to provide a matched and correct minor premise in one step. Our SR-FoT divides the step of proposing a minor premise into two stages: posing the minor premise question (Stage 3) and answering the minor premise question (Stage 4). The task of the "posing the minor premise question" stage is to determine what information about the specific instance in the original question the LLM should acquire to utilize the major premise in answering the "original question". Therefore, in the "posing the minor premise question" stage, besides the guidance and example prompts, the LLM needs access to the "original question," "context," and the "major premise" generated in Stage 2.

Stage 4: \textbf{Minor Premise Production}. 
The task of Stage 4 is to utilize the "context" information provided by the original task, along with the built-in knowledge of the LLM, to answer the minor premise question posed in Stage 3. This stage aims to obtain the correct information about a specific aspect of the particular instance in the original question, leading to the formulation of an accurate and matching minor premise. Given the potential complexity of the minor premise questions, we guide LLM to employ the Chain-of-Thought (CoT) technique to answer the minor premise question and to organize and obtain the minor premise. Furthermore, to avoid the interference caused by excessive information, in this stage, besides the guidance and example prompts, the LLM has access only to the "minor premise question" and "context" without needing to see the "original question" again. The "minor premise question" and "context" already contain all the information necessary for the task of the LLM at this stage; viewing additional information like the "original question" could instead lead to distractions and affect performance. 

Stage 5: \textbf{Final Syllogistic Reasoning}
After the aforementioned stages, complex original knowledge-based reasoning questions can now be answered using syllogistic reasoning. The specific approach involves designing the appropriate task instruction and example prompts, allowing the LLM to engage in syllogistic reasoning based on the major and minor premises generated in earlier stages, to arrive at the answer to the original question. Therefore, in Stage 5, we design the LLM to have access, in addition to the guidance and example prompts, to the "major premise" generated in Stage 2, the "minor premise" generated in Stage 4, and the "original question".

\begin{table*}
  \centering
  \setlength{\tabcolsep}{1pt}
  \begin{tabular}{c|ccc|ccc|ccc}
    \hline
    \multirow{2}{*}{\textbf{Method}} & \multicolumn{3}{c}{GPT-3.5-turbo} & \multicolumn{3}{c}{DeepSeek-V2} & \multicolumn{3}{c}{Qwen1.5-32B-Chat} \\
     & \textbf{ScienceQA} & \textbf{StrategyQA} & \textbf{BoolQ} & \textbf{ScienceQA} & \textbf{StrategyQA} & \textbf{BoolQ} & \textbf{ScienceQA} & \textbf{StrategyQA} & \textbf{BoolQ} \\
    \hline
    \multicolumn{10}{c}{\textit{Single-Round}}\\
    \hline
    Base & 85.9 & 84.8 & 79.4 & 87.0 & 92.3 & 86.3 & 87.4 & 90.8 & 83.4 \\
    CoT~\cite{wei2022chain} & 86.9 & 91.3 & 79.4 & 88.6 & 91.8 & 86.0 & 87.1 & 91.0 & 83.0 \\
    \textbf{SR-FoT (Ours)} & \textbf{87.4} & \textbf{92.1}  & \textbf{81.7} & \textbf{91.8} & \textbf{93.2} & \textbf{87.3} & \textbf{88.7} & \textbf{91.9} & \textbf{84.6} \\
    \hline
    \multicolumn{10}{c}{\textit{Multi-Round}} \\
    \hline
    SC-CoT~\cite{wang2022self} & 87.4 & 91.5 & 82.8 & 88.6 & 92.4 & 87.4 & 87.6 & 92.3 & 84.3 \\
    C-CoT~\cite{fu2022complexity} & 87.4 & 90.2 & 81.0 & 88.5 & 92.3& 87.4 & 87.2 & 92.0 & 83.9 \\
    \textbf{SC-SR-FoT (Ours)} & \textbf{88.9} & \textbf{93.4} & \textbf{85.0} &\textbf{93.0} & \textbf{95.0} & \textbf{89.4} & \textbf{90.4} & \textbf{94.4} & \textbf{88.2}\\
    \hline
  \end{tabular}
  \caption{\label{tab: main-results}
    Comparison with the state-of-the-art methods on the ScienceQA, StategyQA, and BoolQ datasets.
  }
\end{table*}
\section{Experiments}

To evaluate the effectiveness of our SR-FoT, we conducted a series of experiments using both Open-source and closed-source LLMs on several common knowledge-based reasoning question-answer datasets. 
\subsection{Experiment Setup}
\subsubsection{Datasets.}
To fully demonstrate the effectiveness and generalization of our SR-FoT, we conduct a series of experiments on three datasets from different fields.

\textbf{ScienceQA}~\cite{lu2022learn} is a scientific question-answering dataset and contains 21,208 multimodal multiple-choice science questions. It can be divided into three subjects: natural science, language science, and social science. It requires the language model to select one answer from multiple options, usually requiring multi-step reasoning. In our experiment, we employ the test set samples which only have a text context, with a total of 2224. We report the accuracy of our SR-FoT and comparison methods on this set.

\textbf{StrategyQA}~\cite{geva2021did} is a question-answering dataset focusing on open-domain questions. Its questions contain multiple reasoning steps, and a strategy should be used to obtain the answers. In our experiment, we evaluate the methods with accuracy on the train set, which includes 2290 samples.

\textbf{BoolQ}~\cite{clark2019boolq} is a reading comprehension dataset consisting of 16k
samples. They often query for complex, non-factoid information, and require difficult entailment-like inference to solve. In our experiment, we compare the accuracy of our SR-FoT with other methods on the dev set, with a total of 3270.

\subsubsection{Experimental Setting.}

Our experiments are performed using API calls on the proprietary model GPT-3.5-turbo~\cite{ouyang2022training}, the open-source model DeepSeek-v2~\cite{deepseekv2} with 236B parameters, and Qwen1.5-32B-Chat~\cite{qwen1.5} version with 32B parameters. The control group methods include the Base method, Chain of Thought (CoT)~\cite{wei2022chain}, Self-consistency CoT (SC-CoT)~\cite{wang2022self}, and Complexity-based CoT (C-CoT)~\cite{fu2022complexity} methods. Our own approaches included SR-FoT and Self-consistency SR-FoT (SC-SR-FoT), which represents our SR-FoT following the self-consistency sampling and aggregation settings of SC-CoT. In the single-round sampling methods which include Base, CoT and SR-FoT, the hyperparameters on GPT-3.5-turbo and Qwen1.5-32B-Chat are set to top\_p=0.3 and temperature=0.2, while the temperature on DeepSeek-v2 was set to the default recommended value of 1 (DeepSeek only allows the temperature hyperparameter to be adjusted). In the multi-round sampling methods which include SC-CoT, C-CoT and SC-SR-FoT, we perform 10 samplings each. To enhance the diversity of sampling outcomes, the hyperparameters on GPT-3.5-turbo and Qwen1.5-32B-Chat for top\_p and temperature are adjusted to 0.7 and 0.9 respectively. The temperature hyperparameter on DeepSeek remained at the default recommended value of 1. The number of in-context example prompts used in all the methods on the ScienceQA, StrategyQA, and BoolQ datasets are 5, 2, and 2, respectively.

\subsection{Experimental Results and Analyses}
\subsubsection{Performance on ScienceQA.}
Scientific question answering is a task scenario that often requires deductive reasoning. As seen in Tab.~\ref{tab: main-results}, under GPT-3.5-turbo, in the comparison of single-round sampling methods, our SR-FoT outperforms the Base and CoT methods by 1.5\% and 0.5\% respectively and is on par with multi-round sampling methods like SC-CoT and C-CoT. In the comparison of multi-round sampling methods, our SC-SR-FoT exceeds SC-CoT and C-CoT methods by 1.5\%. Under the open-source model DeepSeek-V2, SR-FoT outperforms the Base and CoT by 4.8\% and 3.2\% respectively, even surpassing multi-round sampling methods. What's more, our SC-SR-FoT further increases the accuracy to 93.0\%. Under Qwen1.5-32B-Chat, compared to the Base and CoT methods, our SR-FoT has an improvement of 1.3\% and 1.6\% respectively. Compared to SC-CoT and C-CoT, our SC-SR-FoT also performs better, surpassing them by 2.8\% and 3.2\% respectively. These indicate that our methods achieves greater superiority on the ScienceQA dataset under multiple models.

\subsubsection{Performance on StrategyQA and BoolQ.}
StrategyQA and BoolQ are two other knowledge-based reasoning question-answer datasets that require a true or false judgment based on context or common sense knowledge. From Table 1, for StrategyQA under GPT-3.5-turbo, in the comparison of single-round sampling methods, our SR-FoT outperforms Base and CoT by 7.3\% and 0.8\% respectively; in the comparison of multi-round sampling methods, our SC-SR-FoT exceeds SC-CoT and C-CoT by 1.9\% and 3.2\% respectively. Similar trends are observed under DeepSeek-V2 and Qwen1.5-32B-Chat. In addition, our SR-FoT and SC-SR-FoT also perform the best in both single-round sampling methods and multi-round sampling methods on BoolQ under the three models.

Overall, whether under the closed-source large language model GPT-3.5-turbo or the open-source large language models DeepSeek-V2 and Qwen1.5-32B-Chat, our SR-FoT achieve a superior accuracy compared with other compared methods on the ScienceQA, StrategyQA, and BoolQ datasets. This demonstrates the effectiveness of our SR-FoT. 

It is worth noting that under DeepSeek-V2 and Qwen1.5-32B-Chat, the Base method achieves relatively high results across all three datasets, while the benefits of the CoT method show signs of saturation, and at times perform worse than the Base method. However, our methods, whether under single-round sampling settings (SR-FoT) or multi-round aggregated sampling settings (SC-SR-FoT), are still able to further enhance performance, demonstrating greater potential for performance gains. 
We believe this is because our SR-FoT employs a syllogistic deductive reasoning framework, allowing LLMs to address these knowledge-based reasoning tasks based on a more rigorous reasoning process, thereby achieving better overall performance.

\begin{table*}[h]
  \centering
  \begin{tabular}{c|ccc|cc|c}
    \hline
    \multirow{2}{*}{\textbf{Method}} & \multicolumn{3}{c}{Subject} & \multicolumn{2}{c}{Grade} &\multirow{2}{*}{\textbf{Average}}\\
     & \textbf{Natural} & \textbf{Social} & \textbf{Language} & \textbf{Grade1-6} & \textbf{Grade7-12}& \\
    \hline
    Base &89.4&92.0&84.1&88.2&85.4&87.0\\
    CoT~\cite{wei2022chain}&91.6&96.0&84.8&90.8&85.5&88.6\\
    SC-CoT~\cite{wang2022self} &91.7 &96.8 &84.4 &90.8 &85.3 &88.6 \\
    C-CoT~\cite{fu2022complexity} &91.9 &96.0 &84.7 &90.9 &85.7 &88.5 \\
    w/o explanation &94.2 &\textbf{98.4} &83.2 &90.9 &87.0  &90.4\\
    w/o major premise&93.7 &96.0 &85.1 &91.1 &87.8&90.9 \\
    w/o minor premise&87.3 &94.4 &87.3 &91.0 &83.1&87.7 \\
    all in one stage&91.2 &95.2 &86.0 &91.0 &86.1 &89.8\\
    w/o stage 3&94.0 &96.8 &84.1 &90.1 &88.5 &89.4\\
    \textbf{SR-FoT (Ours)} &95.0 &97.6 &85.7 &91.4 &89.9 &91.8\\
    \textbf{SC-SR-FoT (Ours)} &\textbf{97.1} &96.8 &\textbf{88.4} &\textbf{93.4} &\textbf{92.3} &\textbf{93.0}\\
    \hline
  \end{tabular}
  \caption{\label{tab: ablation3}
     Effectiveness comparisons for subcategories 
 on the ScienceQA dataset with DeepSeek-v2.}
\end{table*}

\subsection{Ablation Study}
\subsubsection{Effectiveness Comparisons for Subcategories.}
As shown in table~\ref{tab: ablation3}, we conduct the experiments on the ScienceQA dataset with DeepSeek-v2. The results demonstrate that our methods can enhance the reasoning performance of the language model across questions of different difficulty levels and various subjects, compared with the state-of-the-art methods. When increasing the consistency of the proposed method, the beneficial effects become more significant. 

\subsubsection{Ablation of Stages.}
As shown in Table~\ref{tab: ablation1}, we conduct experiments on ScienceQA under DeepSeek-V2 to verify the effectiveness of each stage in our method. 
Specifically, 
`all in one stage' denotes using instructions and examples to let the LLM provide the premises based on the question and options, and then directly provide the answers. `w/o stage 3' denotes providing the minor premise directly, instead of posing it as a question first and then answering. The results demonstrate that the completeness of each stage is important. In detail, discarding either the problem explanation or the major and minor premises would decrease the performance. Furthermore, allowing the language model to directly provide the major and minor premise would significantly reduce its performance, demonstrating the necessity of the multi-stage thinking framework in our SR-FoT. 

\begin{table}
  \centering
  \begin{tabular}{c|c}
    \hline
    \textbf{Method} & \textbf{Accuarcy}\\
    \hline
    w/o explanation & 90.4 \\
    w/o major premise& 90.9 \\
    w/o minor premise & 89.8\\
    all in one stage& 87.7\\
    w/o stage 3& 89.4\\
    Ours & \textbf{91.8}\\
    \hline
  \end{tabular}
  \caption{\label{tab: ablation1}
    Ablation study of stages in our proposed methods on the ScienceQA dataset under DeepSeek-V2. 
  }
\end{table}

\begin{table}
  \centering
  \begin{tabular}{c|c}
    \hline
    \textbf{Method} & \textbf{Accuarcy}\\
    \hline
    w/o context in stage 3 &  92.4\\
    add Q\textsubscript{ori} in stage 4 &92.7 \\
    Ours & \textbf{93.2}\\
    \hline
  \end{tabular}
  \caption{\label{tab: ablation2}
   Ablation study of visible information in various stages on the StrategyQA dataset.
  }
\end{table}

\begin{table}
\centering
\setlength{\tabcolsep}{3pt}
  \begin{tabular}{c|c|c|c}
    \hline
     \textbf{Method} & \textbf{Rigorous} & \textbf{Not Rigorous} & \textbf{Rigor Rate} \\
    \hline
    ScienceQA SR-FoT & 41 & 9 & 0.82 \\
    ScienceQA CoT & 38 & 12 & 0.76 \\
    StrategyQA SR-FoT & 46 & 4 & 0.92 \\
    StrategyQA CoT & 44 & 6 & 0.88 \\
    BoolQ SR-FoT & 48 & 2 & 0.96 \\
    BoolQ CoT & 40 & 10 & 0.8 \\
    \hline
  \end{tabular}
  \caption{\label{tab:strictness}
    Fifty cases using CoT and SR-FoT respectively, randomly selected from the three datasets under GPT-3.5-turbo, are analyzed to assess their rigor, and their rigor rates are subsequently calculated.
  }
\end{table}

\begin{table}
  \centering
  \begin{tabular}{c|c|c|c|c}
    \hline
    \textbf{Dataset} & \textbf{MaPE} & \textbf{MiPQE} & \textbf{MiPE} & \textbf{FRPE}\\
    \hline
    ScienceQA & 13 (26\%) & 4 (8\%) & 24 (48\%) & 9 (18\%) \\
    StrategyQA & 15 (30\%) & 5 (10\%) & 14 (28\%) & 16 (32\%) \\
    BoolQ & 10 (20\%) & 14 (28\%) & 10 (20\%) & 16 (32\%)\\
    \hline
  \end{tabular}
  \caption{\label{tab:error}
   Error sources on a random sample of 50 incorrect examples of our SR-FoT from the three datasets using the GPT-3.5-turbo model. `MaPE' denotes major premise error, `MiPQE' denotes minor premise question error, `MiPE' denotes minor premise error and `FRPE' denotes final reasoning process error.
  }
\end{table}

\subsubsection{Impact of Visible Information in Various Stages.}
As shown in Table~\ref{tab: ablation2}, we conduct the experiments on the StrategyQA dataset with DeepSeek-v2. `w/o context in stage 3' denotes that the minor premise question is generated without considering the context. `add Q\textsubscript{ori} in stage 4' denotes that providing the original question, minor premise question, and context all to the LLM during the process of answering the minor premise question. The results demonstrate that both decreasing or increasing the content of the input prompts adversely affect performance. This underlines the appropriateness of the designed visible information at each stage of our SR-FoT.

\subsection{Rigor Analysis}
To more directly analyze whether our SR-FoT improves the rigor of the reasoning process compared to CoT, we randomly select 50 cases from each of the three datasets under GPT-3.5-turbo for manual evaluation. For CoT and SR-FoT, if all intermediate steps from the first step of reasoning to obtaining the final answer are correct and logically progressive, without any factual inconsistencies or self-inconsistencies, we call them rigorous; otherwise, they are not rigorous. The results are in Table~\ref{tab:strictness}. From the table, it can be found that our SR-FoT has a higher rigor rate than CoT on all three datasets, indicating that our SR-FoT has enhanced the rigor and interoperability of LLM reasoning. For specific comparison cases about rigor, please refer to the supplementary materials.

\subsection{Error Analysis}
We randomly selected 50 error cases from each of the three datasets under GPT-3.5-turbo to perform an error analysis of our SR-FoT.
The sources of errors and their respective proportions are as in table~\ref{tab:error}. 
From the error analysis, it can be found that the proportion of different types of errors varies on different datasets. In ScienceQA, most errors stem from the step of extracting suitable minor premises from the question information. In StrategyQA, the main errors stem from the final reasoning process and mistakes in presenting the major premise. In BoolQ, the primary errors originate from the final reasoning process and errors in formulating the minor premise.

\subsection{Case Study}

We give a case on the ScienceQA dataset to show how CoT and SR-FoT work. In Fig.~\ref{nlp_case2}, it can be seen that in the fourth and fifth reasoning steps of CoT, the model misunderstands the rhyme condition and thus infers wrong information, resulting in an incorrect final answer. In SR-FoT, the question explanation points out a reasonable direction for the major premise, then the major premise gives more sufficient rhyme conditions, and the minor premise correctly distinguishes different ending sounds. With their joint help, the model gives the correct final answer. More cases can be found in the supplementary materials.

\section{Conclusion}
In this paper, we have developed a multi-stage syllogistic reasoning framework of thought(SR-FoT) to guide LLMs in solving complex and diverse knowledge-based reasoning question-answering tasks using syllogistic deductive reasoning. 
Experiments across various knowledge-based reasoning datasets under various LLMs demonstrate the effectiveness and advantages of our method.

\bibliography{main}

\end{document}